\documentclass[11pt,onecolumn,peerreview]{IEEEtran}
\usepackage[pdftex]{color,graphicx}
\usepackage{amsfonts,amsmath,amssymb,times,url,latexsym,epsfig,paralist,rotating}
\usepackage{algorithm,algorithmic}
\usepackage[small]{caption}
\usepackage{xspace}
\usepackage{setspace}

\title{Large Scale Language Modeling in Automatic Speech Recognition}
\author{Ciprian Chelba, Dan Bikel, Maria Shugrina, Patrick Nguyen, Shankar Kumar
\thanks{All authors are with Google, Inc., 1600 Amphiteatre Pkwy, Mountain View, CA 94043, USA.}}%

\begin{document}

\maketitle

\singlespacing
\begin{abstract}
Large language models have been proven quite beneficial for a
variety of automatic speech recognition tasks in Google. We
summarize results on Voice Search and a few YouTube speech
transcription tasks to highlight the impact that one can expect from
increasing both the amount of training data, and the size of the language model
estimated from such data. Depending on the task, availability and
amount of training data used, language model size and amount of work
and care put into integrating them in the lattice rescoring step we
observe reductions in word error rate between 6\% and 10\% relative, for
systems on a wide range of operating points between 17\% and 52\%
word error rate.
\end{abstract}
\doublespacing

\section{Introduction}
\label{sec:intro}

A statistical language model estimates the prior probability values $P(W)$ for
strings of words $W$ in a vocabulary ${\cal V}$ whose size is usually in the
tens or hundreds of thousands. Typically the string $W$ is broken
into sentences, or other segments such as utterances in automatic speech
recognition (ASR), which are assumed to be conditionally independent. For
the rest of this chapter, we will assume that $W$ is such a segment,
or sentence. With $W=w_1,w_2,\ldots,w_n$ we get:
\begin{eqnarray}
  \label{intro:bayes}
  P(W)=\prod_{i=1}^nP(w_i|w_1,w_2,\ldots,w_{i-1}) 
\end{eqnarray}

Since the parameter space of $P(w_k|w_1,w_2,\ldots,w_{k-1})$ is too
large, the language model is forced to put the \emph{context}
$W_{k-1}=w_1,w_2,\ldots,w_{k-1}$ into an \emph{equivalence class} determined
by a function $\Phi(W_{k-1})$. As a result,
\begin{equation}
\label{e1}P(W)\cong\prod_{k=1}^nP(w_k|\Phi (W_{k-1})) 
\end{equation}

Research in language modeling consists of finding appropriate
equivalence classifiers $\Phi$ and methods to estimate
$P(w_k|\Phi(W_{k-1}))$.

The most successful paradigm in language modeling uses the
\emph{$(n-1)$-gram} equivalence classification, that is, defines%
$$
\Phi (W_{k-1})\doteq w_{k-n+1},w_{k-n+2},\ldots,w_{k-1}
$$
Once the form $\Phi (W_{k-1})$ is specified, only the problem of
estimating $P(w_k|\Phi (W_{k-1}))$ from training data remains.
In most practical cases, $n=3$ which leads to a \emph{trigram}
language model.

A commonly used quality measure for a given model $M$ is related to
the entropy of the underlying source and was introduced under
the name of \emph{perplexity} (PPL)~\cite{jelinek97}:
\begin{eqnarray}
\label{basic_lm:ppl}
PPL(M) = exp(-\frac{1}{N} \sum_{k=1}^{N}\ln{[P_M(w_k|W_{k-1})]}) 
\end{eqnarray}

A more relevant metric for ASR is the word error rate (WER) achieved when
using a give language model in a speech recognition system.

The distributed language model architecture described
in~\cite{brants-EtAl:2007:EMNLP-CoNLL2007} can be used for training
and serving very large language models.
We have implemented lattice rescoring in this setup, and experimented
with such large distributed language models on various Google internal
tasks.

\section{Voice Search Experiments}
\label{sec:vs_wer}

We have trained query LMs in the following setup~\cite{Chelba2010}:
\begin{itemize}
\item vocabulary size: 1M words, OOV rate 0.57\%
\item training data: 230B words, a random sample of anonymized
  queries from google.com that did not trigger spelling correction.
\end{itemize}
The test set was gathered using an Adroid application. People were
prompted to speak a set of random google.com queries selected from a time
period that does not overlap with the training data.

\begin{figure}[t]
  \centering
  \includegraphics[width=0.9\columnwidth]{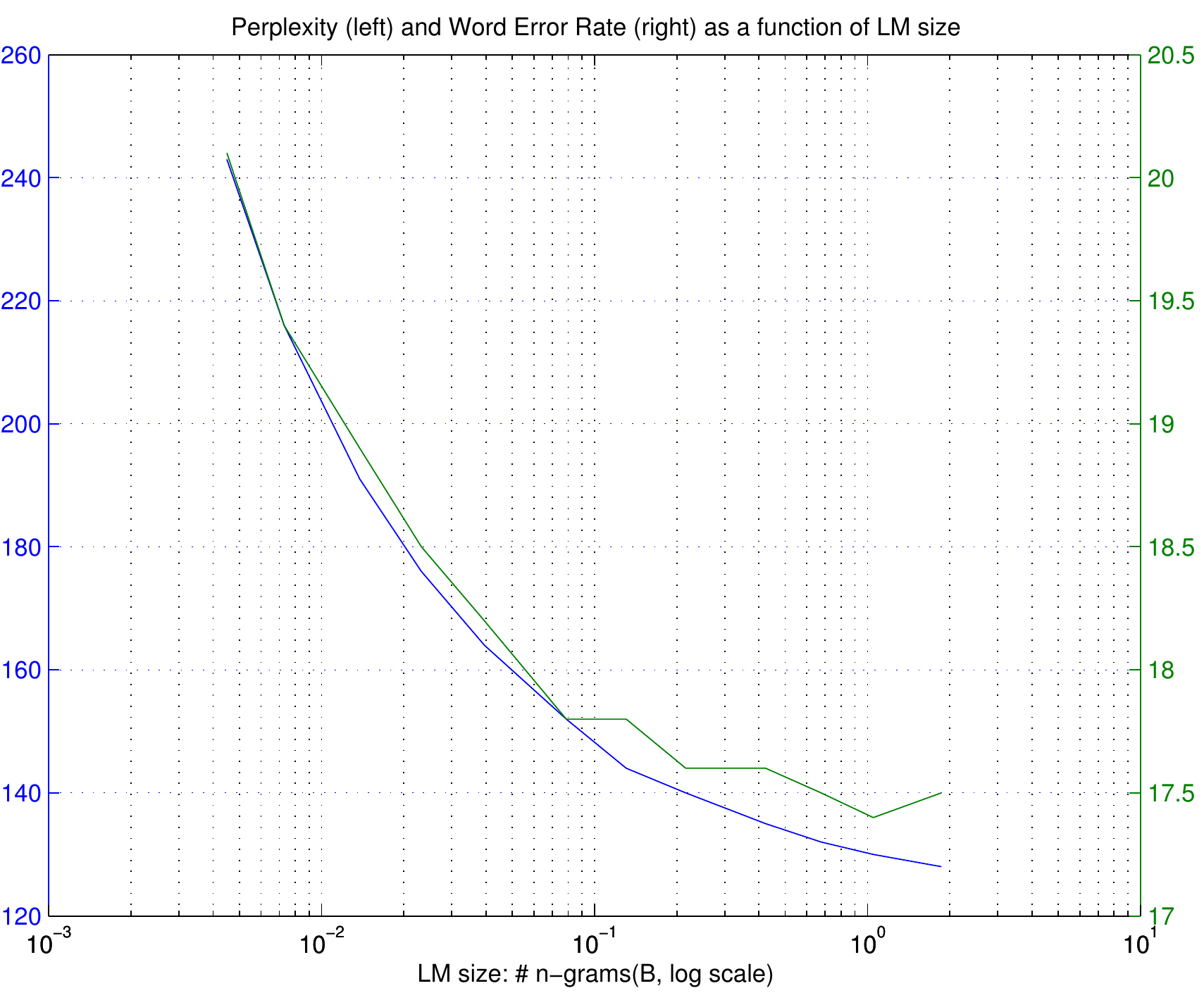}
  \caption{3-gram language model perplexity and word error rate as a
    function of language model size; lower curve is PPL.}
  \label{fig:ppl_wer_3gram}
\end{figure}

\begin{figure}[t]
  \centering
  \includegraphics[width=0.9\columnwidth]{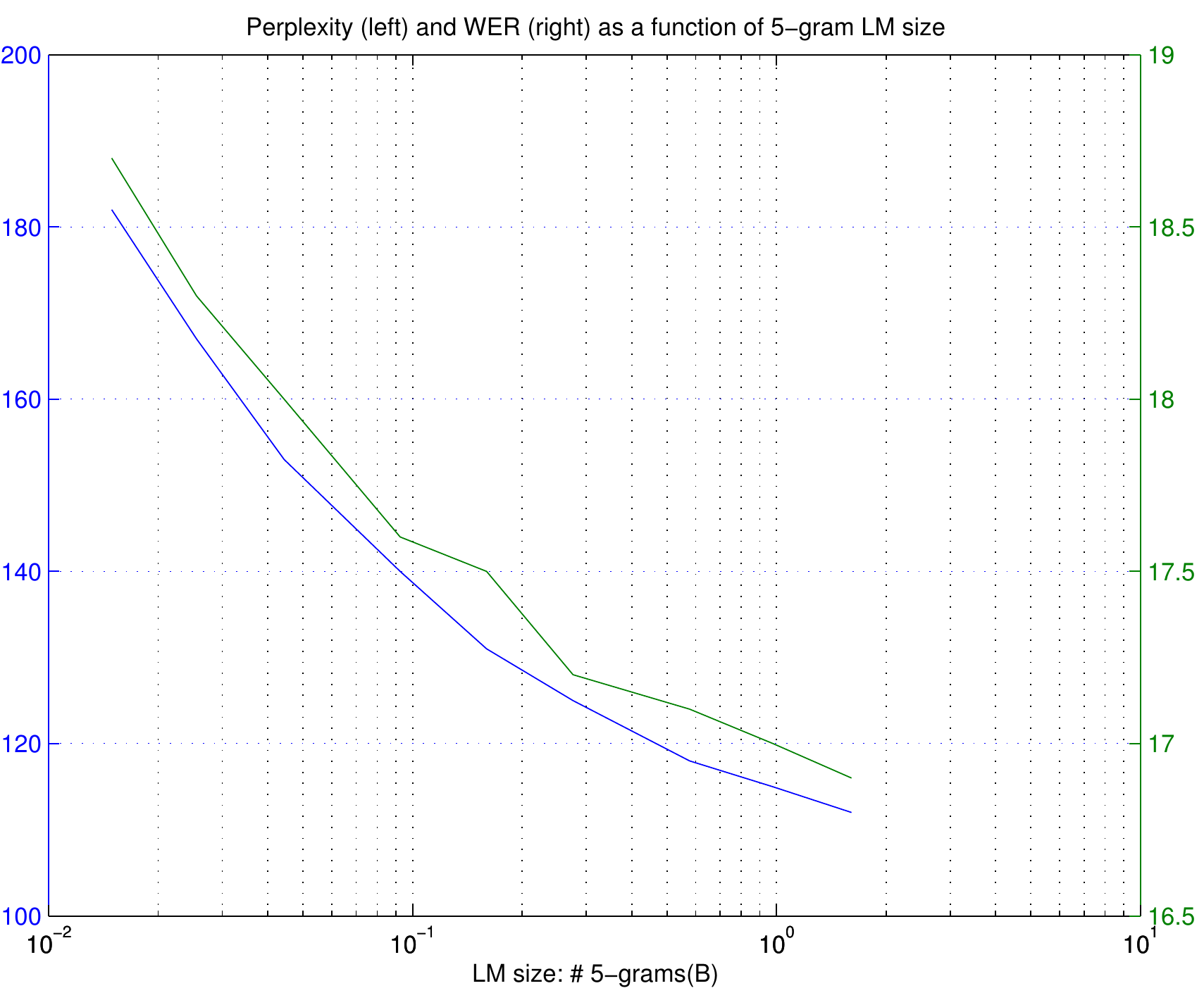}
  \caption{5-gram language model perplexity and word error rate as a
    function of language model size; lower curve is PPL.}
  \label{fig:ppl_wer_5gram}
\end{figure}

The work described in~\cite{harb09}~and~\cite{riley09} enables us to
evaluate relatively large query language models in the 1-st pass of
our ASR decoder by representing the language model in the
OpenFst~\cite{openfst}
framework. Figures~\ref{fig:ppl_wer_3gram}-\ref{fig:ppl_wer_5gram}
show the PPL and word error rate (WER) for two language models (3-gram
and 5-gram, respectively) built on the 230B training data, after
entropy pruning to various sizes in the range 15 million - 1.5 billion
n-grams.

As can be seen, perplexity is very well correlated with WER, and the
size of the language model has a significant impact on speech
recognition accuracy: increasing the model size by two orders of
magnitude reduces the WER by 10\% relative.

We have also implemented lattice rescoring using the distributed language
model architecture described
in~\cite{brants-EtAl:2007:EMNLP-CoNLL2007}, see the results presented
in Table~\ref{lattice_rescoring}. 

This enables us to validate empirically the fact that rescoring
lattices generated with a relatively small 1-st pass language model
(in this case 15 million 3-gram, denoted 15M 3-gram in
Table~\ref{lattice_rescoring}) yields the same results as 1-st pass
decoding with a large language model.  A secondary benefit of the
lattice rescoring setup is that one can evaluate the ASR performance
of much larger language models.

\begin{table}[h]
  \begin{center}
    \begin{tabular}{llrrr}
      Pass  & Language Model                & Size  & PPL & WER (\%) \\\hline\hline
      1st   & 15M 3-gram                    & --- & 191 & 18.7 \\
      1st   & 1.6B 5-gram         & LARGE, pruned & 112 & 16.9 \\\hline
      2nd   & 15M 3-gram                    & --- & 191 & 18.8 \\
      2nd   & 1.6B 5-gram         & LARGE, pruned & 112 & 16.9 \\
      2nd   & 12.7B 5-gram        & LARGE         & 108 & 16.8 \\\hline
    \end{tabular}
  \end{center}
  \caption{\label{lattice_rescoring} Speech recognition language model
    performance when used in the 1-st pass or in the 2-nd
    pass---lattice rescoring.}
\end{table}

\section{YouTube Experiments}
\label{sec:yt_wer}

YouTube data is extremely challenging for current ASR
technology. As far as language modeling is concerned, the variety of topics and
speaking styles makes a language model built from a web crawl a very
attractive choice.

\subsection{2011 YouTube Test  Set}

A second batch of experiments were carried out in a different training
and test setup, using more recent and also more challenging YouTube
speech data.

On the acoustic modeling side, the training data for the YouTube
system consisted of approximately 1400 hours of data from YouTube. The
system used 9-frame MFCCs that were transformed by LDA and SAT was
performed. Decision tree clustering was used to obtain 17552 triphone
states, and STCs were used in the GMMs to model the features. The
acoustic models were further improved with bMMI~\cite{bMMI}. During
decoding, Constrained Maximum Likelihood Linear Regression (CMLLR) and
Maximum Likelihood Linear Regression (MLLR) transforms were applied.

The training data used for language modeling consisted of
Broadcast news acoustic transcriptions (approx.\ 1.6 million words),
Broadcast news LM text distributed by LDC (approx.\ 128 million
words), and a web crawl from October 2008 (approx.\ 12 billion
words). Each data source was used to train a separate interpolated
Kneser-Ney 4-gram language model, of size 3.5 million, 112 million and
5.6 billion n-grams, respectively.

The first pass language model was obtained by interpolating the three
components above, after pruning each of them to 3-gram order
and about 10M n-grams. Interpolation weights were
estimated such that they maximized the probability of a held-out set
consisting of manual transcription of YouTube utterances.

For lattice rescoring, the three language models were combined with the 1-st
pass acoustic model score and the insertion penalty using
MERT~\cite{lattice_mert}.

The test set consisted of 10 hours of randomly selected YouTube speech
data.

Table~\ref{yt2011_lattice_rescoring} presents the results in various
rescoring configurations:
\begin{itemize}
\item \emph{2nd, MERT} uses lattice MERT to compute the optimal weights for
  mixing the three language model scores, along with acoustic model
  score and insertion penalty. It achieves 3.2\% absolute reduction in
  WER. Despite the very high error rate of the baseline this amounts to 6\%
  relative reduction in WER.
\item \emph{2nd, unif} uses uniform weights across the three language
  models, quantifying the gain that can be attributed to MERT (0.6\%
  absolute).
\item \emph{2nd, no www} throws away the www LM from the mix to evaluate
  its contribution: 1.2\% absolute reduction in WER.
\end{itemize}

\begin{table}[h]
  \begin{center}
    \begin{tabular}{llrr}
      Pass             & Language Model        & Size & WER (\%) \\\hline\hline
      1st              & 14M 3-gram            & ---  & 54.4 \\
      2nd, MERT        & 5.6B 4-gram           & LARGE& 51.2 \\
      2nd, unif        & 5.6B 4-gram           & LARGE& 51.8 \\
      2nd, no www LM   & 112M 4-gram           & ---  & 53.0 \\\hline
    \end{tabular}
  \end{center}
  \caption{\label{yt2011_lattice_rescoring} YouTube 2011 test set:
    Lattice rescoring using a large language model trained on web crawl.}
\end{table}

Experiments on a development set collected at the same time with the
test set insert the large LM rescoring at various stages in the
rescoring pipeline, using increasingly powerful acoustic models, as
reported in~\cite{jaitly:interspeech2012}. The
results are reported in Table~\ref{yt2011_dev_lattice_rescoring}.
\begin{table}[h]
  \begin{center}
    \begin{tabular}{llrrr}
      Pass           & Acoustic Model        & Language Model & Size & WER (\%) \\\hline\hline
      Baseline       & baseline AM           & 14M 3-gram     & ---  & 52.8 \\
      2nd            & baseline AM           & 5.6B 4-gram    & LARGE& 49.4 \\\hline
      better AM      & DBN + tuning          & 14M 3-gram     & ---  & 49.4 \\
      2nd            & DBN + tuning          & 5.6B 4-gram    & LARGE& 45.4 \\\hline
      even better AM & MMI DBN + tuning      & 14M 3-gram     & ---  & 48.8 \\
      2nd            & MMI DBN + tuning      & 5.6B 4-gram    & LARGE& 45.2 \\\hline
    \end{tabular}
  \end{center}
  \caption{\label{yt2011_dev_lattice_rescoring} YouTube 2011 dev set:
    Lattice rescoring using a large language model trained on web
    crawl. Lattices are generated with increasingly powerful acoustic
    models.}
\end{table}

We observe consistent gains between 6\% and 9\% relative, 3.4-4.0\%
absolute at various operating points in WER due to more powerful
acoustic models. As a side note, the gains from large LM rescoring are
comparable to those obtained by using deep-belief NN acoustic models (DBN).

\subsection{2008 YouTube Test Set}

In a different batch of YouTube experiments, Thadani et
al.~\cite{bikel:interspeech2012} train a language model on a web crawl
from 2010, filtered to retain only documents in English.

The training data used for language modeling consisted of
Broadcast news acoustic transcriptions (approx.\ 1.6 million words), Broadcast news LM text
distributed by LDC (approx.\ 128 million words), and a web crawl from
2010 (approx.\ 59 billion words). Each data source was used to
train a separate interpolated Kneser-Ney 4-gram language model, of size 3.5 million, 112
million and 19 billion n-grams, respectively.

The first pass language model was obtained by interpolating the three
components above, after pruning each of them to 3-gram order
and about 10M n-grams.

For lattice rescoring, the three unpruned language models were combined using
linear interpolation.

For both first-pass and rescoring language models, interpolation
weights were estimated such that they maximized the probability of a
held-out set consisting of manual transcription of YouTube utterances.

The test corpus consisted of 77 videos containing news broadcast style
material downloaded in 2008~\cite{Alberti2009}. They were automatically segmented into
short utterances based on pauses between speech. The audio was
transcribed at high quality by humans trained in the task.

Table~\ref{yt2008_lattice_rescoring} highlights the large LM rescoring
results presented in \cite{bikel:interspeech2012}.
\begin{table}[h]
  \begin{center}
    \begin{tabular}{llrrr}
      Pass  & Language Model        & Size  & WER (\%) \\\hline\hline
      1st   & 14M 3-gram            & ---   & 34.6 \\
      2nd   & 19B 4-gram            & LARGE & 31.8 \\\hline
    \end{tabular}
  \end{center}
  \caption{\label{yt2008_lattice_rescoring} YouTube 2008 test set:
    Lattice rescoring using a large language model trained on web
    crawl.}
\end{table}

The large language model used for lattice rescoring decreased the WER
by 2.8\% absolute, or 8\% relative, a significant improvement in
accuracy. \footnote{Unlike the Voice Search experiments reported in
  Table~\ref{lattice_rescoring}, no interpolation between the first
  and the second pass language model was performed. In our experience
  that consistently yields small gains in accuracy.}

\section{Conclusions}
\label{sec:conclusions}

Large n-gram language models are a simple yet very effective way of improving the
performance of real world ASR systems. Depending on the task,
availability and amount of training data used, language model size and
amount of work and care put into integrating them in the lattice
rescoring step we observe improvements in WER between 6\% and 10\%
relative.

\newpage
\bibliographystyle{IEEEbib}
\bibliography{lm}

\begin{thebibliography}{10}

\bibitem{jelinek97}
Frederick Jelinek,
\newblock {\em Information Extraction From Speech And Text}, chapter~8, pp.
  141--142,
\newblock MIT Press, 1997.

\bibitem{brants-EtAl:2007:EMNLP-CoNLL2007}
T.~Brants, A.~C. Popat, P.~Xu, F.~J. Och, and J.~Dean,
\newblock ``Large language models in machine translation,''
\newblock in {\em Proceedings of the 2007 Joint Conference on Empirical Methods
  in Natural Language Processing and Computational Natural Language Learning
  (EMNLP-CoNLL)}, 2007, pp. 858--867.

\bibitem{Chelba2010}
C.~Chelba, J.~Schalkwyk, T.~Brants, V.~Ha, B.~Harb, W.~Neveitt, C.~Parada, and
  P.~Xu,
\newblock ``Query language modeling for voice search,''
\newblock in {\em Proc. of SLT}, 2010.

\bibitem{harb09}
B.~Harb, C.~Chelba, J.~Dean, and S.~Ghemawat,
\newblock ``Back-off language model compression,''
\newblock in {\em Proceedings of Interspeech}, Brighton, UK, 2009, ISCA, pp.
  325--355.

\bibitem{riley09}
C.~Allauzen, J.~Schalkwyk, and M.~Riley,
\newblock ``A generalized composition algorithm for weighted finite-state
  transducers,''
\newblock in {\em Proc. Interspeech}, 2009, pp. 1203--1206.

\bibitem{openfst}
C.~Allauzen, M.~Riley, J.~Schalkwyk, W.~Skut, and M.~Mohri,
\newblock ``Open{F}st: A general and efficient weighted finite-state transducer
  library,''
\newblock in {\em Proceedings of the Ninth International Conference on
  Implementation and Application of Automata, (CIAA 2007)}. 2007, vol. 4783 of
  {\em Lecture Notes in Computer Science}, pp. 11--23, Springer,
\newblock {\tt http://www.openfst.org}.

\bibitem{bMMI}
D.~Povey, D.~Kanevsky, B.~Kingsbury, B.~Ramabhadran, G.~Saon, and
  K.~Visweswariah,
\newblock ``Boosted {MMI} for model and feature space discriminative
  training,''
\newblock in {\em Proceedings of ICASSP}, April 2008, pp. 4057 --4060.

\bibitem{lattice_mert}
W.~Macherey, F.~Och, I.~Thayer, and J.~Uszkoreit,
\newblock ``Lattice-based minimum error rate training for statistical machine
  translation,''
\newblock in {\em Proceedings of the 2008 Conference on Empirical Methods in
  Natural Language Processing (EMNLP)}, 2008, pp. 725--734.

\bibitem{jaitly:interspeech2012}
D.~Jaitly, P.~Nguyen, A.~Senior, and V.~Vanhoucke,
\newblock ``Application of pretrained deep neural networks to large vocabulary
  speech recognition,''
\newblock in {\em Proceedings of Interspeech}, 2012.

\bibitem{bikel:interspeech2012}
K.~Thadani, F.~Biadsy, and D.~Bikel,
\newblock ``On-the-fly topic adaptation for youtube video transcription,''
\newblock in {\em Proceedings of Interspeech}, 2012.

\bibitem{Alberti2009}
C.~Alberti, M.~Bacchiani, A.~Bezman, C.~Chelba, A.~Drofa, H.~Liao, P.~Moreno,
  T.~Power, A.~Sahuguet, M.~Shugrina, and O.~Siohan,
\newblock ``An audio indexing system for election video material,''
\newblock in {\em Proceedings of ICASSP}, 2009, pp. 4873--4876.

\end{thebibliography}

\end{document}